\documentclass[letterpaper, 10pt, twocolumn]{article}

\usepackage[a4paper, left=0.8in, right=0.8in, top=1in, bottom=1in, columnsep=0.4in]{geometry} 
\usepackage{amsmath, bm} 
\usepackage{graphicx} 
\usepackage{dcolumn} 
\usepackage{booktabs} 
\usepackage{multirow} 
\usepackage{subcaption} 
\usepackage{caption} 
\usepackage{titlesec} 
\usepackage{fancyhdr} 
\usepackage{libertine} 
\usepackage{hyperref} 
\usepackage{tcolorbox} 
\usepackage{enumitem} 
\usepackage{color,soul}

\hypersetup{
    colorlinks=true,
    linkcolor=black,
    urlcolor=black,
    citecolor=black
}

\titleformat{\section}{\Large\bfseries\rmfamily}{\thesection}{1em}{}
\titleformat{\subsection}{\large\bfseries\rmfamily}{\thesubsection}{1em}{}
\titlespacing*{\section}{0pt}{1.5em}{1em} 
\titlespacing*{\subsection}{0pt}{1em}{0.5em} 

\title{\textbf{Dataset for Real-World Human Action Detection Using FMCW mmWave Radar}}
\author{
    Dylan Jayabahu$^{1}$, Parthipan Siva$^{1,2}$ \\
    $^1$Chirp Inc.\\
    $^2$Vision and Image Processing Group, Systems Design Engineering, University of Waterloo\\
    \texttt{dylanjayabahu@gmail.com},
    \texttt{parthipan.siva@mychirp.com}
}
\date{}

\begin{document}

\maketitle


\begin{tcolorbox}[colback=gray!10, sharp corners, boxrule=0pt, left=5pt, right=5pt, top=5pt, bottom=5pt]
\begin{abstract}
Human action detection using privacy-preserving mmWave radar sensors is studied for its applications in healthcare and home automation. Unlike existing research, limited to simulations in controlled environments, we present a real-world mmWave radar dataset with baseline results for human action detection.
\end{abstract}
\end{tcolorbox}

\section{Introduction}

Millimeter-wave (mmWave) radar is rapidly emerging as a privacy-preserving sensing technology for home monitoring, with applications ranging from home automation to healthcare. As our population ages and more older adults prefer to age in place—driven by both personal choice and limited availability in care facilities—technology becomes crucial to scale in-home care. For mmWave sensors to effectively support in-home monitoring, they must accurately track individuals and recognize key actions like falls and transfers.

While prior studies have explored human action recognition (HAR) using mmWave radar, they are limited to controlled environments with actors performing a set of predefined actions \cite{wuDTXTYTZTPaper,jin2019multiple,ullmann2023survey}. In this paper, we present a dataset collected in real homes, capturing natural activities across 28 residences; among this data we focus on transfer actions in and out of chairs. We evaluate the performance of an existing baseline CNN model for action detection in our real-world dataset to illustrate limitations and challenges.

\section{Data Collection}

In this section, we detail the data collection methods used to build a robust, real-world dataset for human action detection. 


\subsection{Sensor}

\begin{figure}[t]
    \centering
    \includegraphics[width=0.24\columnwidth]{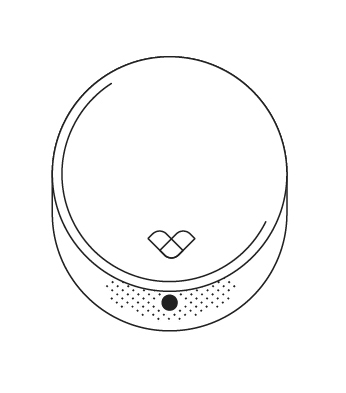}
    \includegraphics[width=0.74\columnwidth]{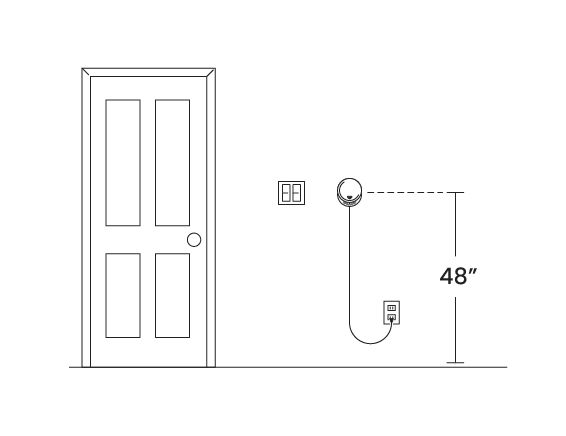}
    \caption{Chirp Smart Home Sensor (left) is installed in the homes at 122 cm (48") height (right).}
    \label{fig:chirpsensor}
\end{figure}

\begin{figure}[t]
    \centering
    \includegraphics[width=0.95\columnwidth]{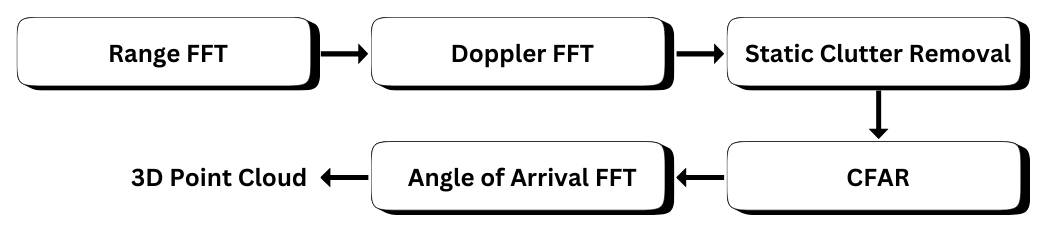}
    \caption{Radar signal processing to obtain 3D point cloud.}
    \label{fig:radarsignalprocessing}
\end{figure}

Data was collected using the Chirp Smart Home Sensor \cite{mychirp}, which incorporates the TI IWR6843AOP FMCW mmWave sensor (Figure \ref{fig:chirpsensor}). To reduce data storage and bandwidth issues, the sensor captures sparse 3D point cloud data at 10 frames per second (FPS) through a radar processing chain illustrated in Figure \ref{fig:radarsignalprocessing}. Each point in the 3D point cloud includes coordinates (x, y, z), velocity, and signal-to-noise ratio (SNR). Using this radar point cloud data, Doppler-time, range-time, azimuth-time, elevation-time, and time-based images along each axis (x, y, z) can be generated as per the approach presented in \cite{wuDTXTYTZTPaper}.

\begin{figure}[h!]
    \centering
    \begin{subfigure}{0.2\textwidth}
        \centering
        \includegraphics[width=\textwidth]{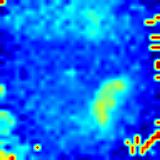} 
        \caption{Frame 1}
        \label{fig:subplot1}
    \end{subfigure}%
    \hspace{0.01\textwidth} 
    \begin{subfigure}{0.2\textwidth}
        \centering
        \includegraphics[width=\textwidth]{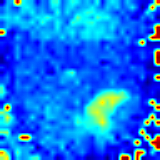} 
        \caption{Frame 6}
        \label{fig:subplot2}
    \end{subfigure} \\
    \begin{subfigure}{0.2\textwidth}
        \centering
        \includegraphics[width=\textwidth]{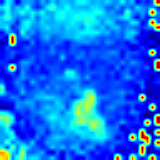} 
        \caption{Frame 11}
        \label{fig:subplot3}
    \end{subfigure}%
    \hspace{0.01\textwidth} 
    \begin{subfigure}{0.2\textwidth}
        \centering
        \includegraphics[width=\textwidth]{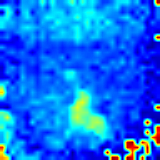} 
        \caption{Frame 16}
        \label{fig:subplot4}
    \end{subfigure}
    \caption{Low resolution thermopile data used for annotation.}
    \label{fig:thermal}
\end{figure}

To support radar data annotation while maintaining privacy, we supplemented the mmWave sensor with a low-resolution 32x32 pixel thermopile sensor, capturing thermal video at 3 FPS. Figure \ref{fig:thermal} illustrates sample actions as viewed by the thermal sensor. While the sparsity of the radar data alone complicates action annotation, the thermal sensor offers enough visual cues to accurately identify actions while ensuring individual privacy.

\subsection{Collection Sites}

Data was collected from 28 unique homes, each occupied by a single older adult participant. Collection occurred over a continuous 24-hour period, allowing participants to follow their usual routines. Due to visitors, actions present within a single home may involve multiple individuals. However, because participants were independently recruited, same individual will not appear in multiple homes.

Participants were given two to three devices and instructed to install them in common living areas, such as kitchens and living rooms. In total, data was collected across the following room types:

\begin{itemize}
\setlength\itemsep{0em} 
\setlength\parskip{0em} 
    \item Kitchens/Dining Rooms: 20
    \item Living Rooms: 24
    \item Multipurpose Rooms: 2
    \item Bathrooms: 2
\end{itemize}

Devices were generally installed at a height of 120 cm from the ground, roughly the standard height of a light switch, and placed to face areas where occupants commonly interacted (Figure \ref{fig:chirpsensor}). Installation height varied based on wall space availability and participant preferences.

\subsection{Data annotation}

Action annotation was achieved by manually reviewing the thermal video, with a focus on two key actions: sitting down and standing up. These actions correspond to transfers into and out of various types of furniture, such as chairs, stools, sofas, and toilets. Sit-down and stand-up actions were selected for their relevance to mobility assessment and their frequent use in clinical evaluations, such as the Timed Up and Go (TUG) test and the 30-Second Sit-to-Stand test.

This dataset thus captures common daily activities, enabling the development of mmWave-based HAR models capable of recognizing key actions that are critical to assessing and supporting mobility in home settings.

\subsection{Dataset}

\begin{figure}[t]
    \centering
    \includegraphics[width=0.8\columnwidth]{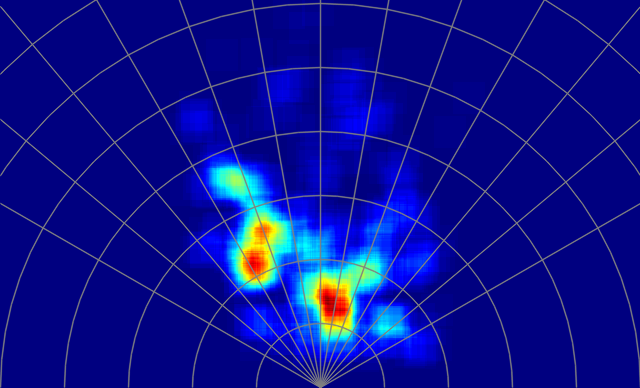}
    \caption{Distribution of sit-down and stand-up action locations in the dataset. Each arc is one meter apart.}
    \label{fig:actiondist}
\end{figure}

Across 28 homes, a total of 458 sit-down and 454 stand-up actions were annotated within the 48 rooms. The spatial distribution of these actions is presented in Figure \ref{fig:actiondist}. For a robust evaluation process, the data was divided such that 3 homes were selected for testing, 3 for validation, and the remaining 22 were designated for training.

Non-action data was generated using a 5-second sliding window on each recording, ensuring a 0\% overlap between windows and with annotated action locations. Any window not meeting the minimum required radar points was removed. The resulting data distribution is detailed in Table \ref{tab:datasetStats}.

\begin{table}[th]
    \caption{Dataset split statistics}
    \centering
    \begin{tabular}{lcccc}
        \toprule
         & Test & Valid & Train & Train + Aug \\
        \midrule
        \# of Homes & 3 & 3 & 22 & 22 \\
        \# of Rooms & 6 & 6 & 36 & 36 \\
        sit-down & 65 & 60 & 333 & 7659 \\
        stand-up & 62 & 58 & 334 & 7682 \\
        other & N/A & 6,666 & 29,803 & 29,803 \\
        \bottomrule
    \end{tabular}
    \label{tab:datasetStats}
\end{table}

\begin{figure*}[ht]
    \centering
    \includegraphics[width=0.8\textwidth]{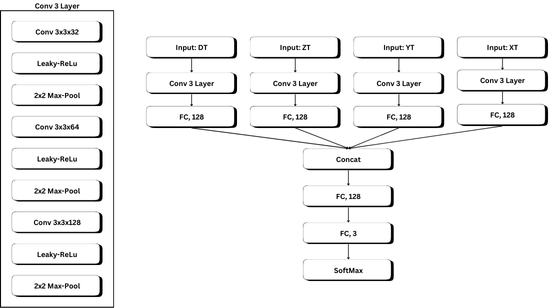}
    \caption{Baseline CNN model architecture from \cite{wuDTXTYTZTPaper,jin2019multiple}.}
    \label{model}
\end{figure*}

An imbalance exists between non-action and action classes. To address this, action windows were over sampled with various augmentations applied to the 3D point cloud data. These random augmentations include:
\begin{itemize} 
\setlength\itemsep{0em} 
\setlength\parskip{0em} 
\item Scaling of signal-to-noise ratio (SNR) 
\item Scaling position along the radial axis 
\item Shifting along the x, y, and z axes 
\item Reflection along the x-axis 
\item Rotation around the z-axis 
\item Random drops of data points
\end{itemize}
This oversampling with augmentations produces a more balanced dataset, as shown in Table \ref{tab:datasetStats}.

\section{Experiment Setup}

We use Doppler-time (DT), x-time (XT), y-time (YT), and z-time (ZT) data images as inputs to our CNN models. Each feature image is generated from the 3D point cloud data following the procedure outlined in \cite{wuDTXTYTZTPaper}.

Unlike existing methods that emphasize classification accuracy within a predefined set of actions \cite{10018429,kruse2023radar}, our testing employs a sliding window detection approach. This approach better reflects real-world action detection scenarios, where ``non-action'' encompasses all other possible actions in a typical home environment.

We assess detection performance using standard metrics: Recall, Precision, and F1-Score for each class.

\section{Detection Model}

As a baseline model we use the 3-layer CNN architecture of \cite{jin2019multiple,wuDTXTYTZTPaper}. The architecture is illustrated in Figure~\ref{model}. To evaluate baseline performance we study the effect of DT, DT+ZT, and DT+ZT+YT+XT as inputs to the model.

Adam optimizer and cross entropy loss was used to train the network from scratch for up to 25 epoch. Early stopping was used based on maximizing the F2-Score on validation dataset. F2 chosen to allow recall to be weighted higher due to the class imbalance problem.

\section{Results \& Discussion}

The detection confusion matrix is presented in Table~\ref{tab:cf1}--~\ref{tab:cf3}. Overall recall and precision are presented in Table~\ref{tab:pr}

\begin{table}[h!]
    \centering
    \begin{minipage}{0.3\textwidth}
        \begin{tabular}{|c|c|c|c|}
            \multicolumn{1}{c}{} & \multicolumn{3}{c}{Predicted} \\ \cline{2-4}
             \multicolumn{1}{c|}{}& Other & Sit-Down & Stand-Up \\ \hline
             Other & N/A & 270 & 461 \\ \hline
            Sit-Down & 33 & 32 & 0 \\ \hline
            Stand-Up & 31 & 0 & 31 \\ \hline
        \end{tabular}
        \caption{DT Results}
        \label{tab:cf1}
    \end{minipage}%
    \hspace{0.1\textwidth} 
    \begin{minipage}{0.3\textwidth}
        \begin{tabular}{|c|c|c|c|}
             \multicolumn{1}{c}{} & \multicolumn{3}{c}{Predicted} \\ \cline{2-4}
             \multicolumn{1}{c|}{}& Other & Sit-Down & Stand-Up \\ \hline
             Other & N/A & 17 & 34 \\ \hline
            Sit-Down & 35 & 30 & 0 \\ \hline
            Stand-Up & 50 & 0 & 12 \\ \hline
        \end{tabular}
        \caption{DT+ZT Results}
        \label{tab:cf2}
    \end{minipage}%
    \hspace{0.1\textwidth} 
    \begin{minipage}{0.3\textwidth}
        \begin{tabular}{|c|c|c|c|}
             \multicolumn{1}{c}{} & \multicolumn{3}{c}{Predicted} \\ \cline{2-4}
             \multicolumn{1}{c|}{}& Other & Sit-Down & Stand-Up \\ \hline
             Other & N/A & 390 & 77 \\ \hline
            Sit-Down & 12 & 53 & 0 \\ \hline
            Stand-Up & 45 & 0 & 17 \\ \hline
        \end{tabular}
        \caption{DT+ZT+YT+XT Results}
        \label{tab:cf3}
    \end{minipage}
\end{table}

\begin{table}[h!]
    \centering
    \small
    \begin{tabular}{|c|c|c|c|c|}
        \hline
        \multirow{2}{*}{} & \multicolumn{2}{c|}{Validate} & \multicolumn{2}{c|}{Test} \\ \cline{2-5}
         & Recall & Precision & Recall & Precision \\ \hline
         DT & 0.729 & 0.079 & 0.496 & 0.074  \\ \hline
         DT+ZT & 0.746 & 0.427 & 0.331 & 0.452 \\ \hline
         DT+ZT+YT+XT & 0.873 & 0.106 & 0.551 & 0.130 \\ \hline
    \end{tabular}
    \caption{Overall recall and precision for detecting both actions.}
    \label{tab:pr}
\end{table}

\begin{figure}[h!]
    \centering
    \begin{subfigure}{0.22\textwidth}
        \centering
        \includegraphics[width=\textwidth]{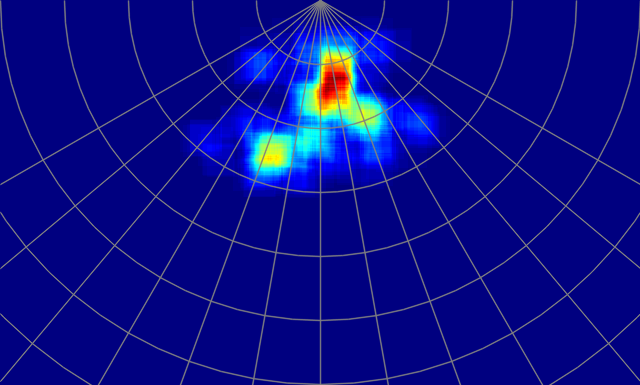} 
        \caption{Train}
        \label{fig:image1}
    \end{subfigure}%
    \hspace{0.01\textwidth} 
    \begin{subfigure}{0.22\textwidth}
        \centering
        \includegraphics[width=\textwidth]{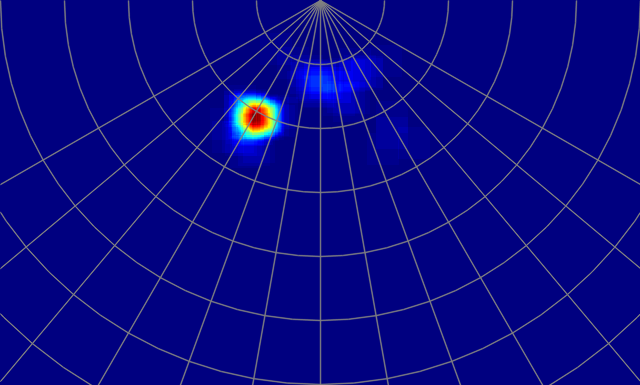} 
        \caption{Valid}
        \label{fig:image2}
    \end{subfigure}%
    \hspace{0.01\textwidth} 
    \begin{subfigure}{0.22\textwidth}
        \centering
        \includegraphics[width=\textwidth]{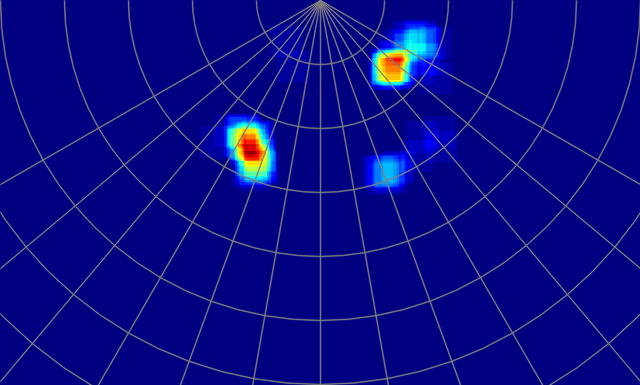} 
        \caption{Test}
        \label{fig:image3}
    \end{subfigure}
    \caption{Distribution of action location by dataset splits.}
    \label{fig:three_images}
\end{figure}

While test results remain poor, validation recall results perform comparatively better, suggesting significant variability between the training, validation, and test sets. This variability arises not only from differences in individuals performing the actions but also from distinct action locations within the environment. Although the dataset includes approximately 450 instances of sit-down and stand-up actions, there are only 48 distinct rooms, limiting the variety of spatial locations and orientations where these actions occur. This limitation is evident when examining the spatial distribution of action locations across the dataset splits (Figure~\ref{fig:three_images}); actions in the test set frequently occur in locations not represented in the training or validation sets.

Precision rates remain low, indicating high variation in the non-action class and prioritization of recall in early stopping. Despite using class weights and oversampling, an optimal balance between recall and precision wasn't achieved.

Comparing DT (F1=0.137), DT+ZT (F1=0.382), and DT+ZT+YT+XT (F1=0.211), we observe that DT+ZT achieves a better balance of recall and precision. This is intuitive, as Z-axis variation for sit-down and stand-up actions remains consistent regardless of location within the room, while YT and XT vary by location. Additionally, ZT enhances the Doppler (speed) context provided by DT only.

\section{Conclusions}

We present a robust real-world dataset for action detection using mmWave radar in natural environments. Unlike existing datasets that simulate actions of interest, our dataset captures real-world activity in the homes of older adults and annotates actions within the captured data. This results in a more realistic dataset; however, it also leads to higher variance in both the environment and the actions performed. We demonstrate that, under these challenging conditions, existing baseline approaches perform poorly.

\bibliographystyle{IEEEtran}
\bibliography{references.bib}

\end{document}